%% file: main.tex
\documentclass{article}

\usepackage{microtype}
\usepackage{graphicx}
\usepackage{subcaption}
\usepackage{booktabs} %
\usepackage{xcolor}
\usepackage[table]{xcolor}
\usepackage{array}
\usepackage[most]{tcolorbox}
\usepackage{fvextra}

\usepackage{hyperref}

\definecolor{lightgreen}{RGB}{0,110,0} 
\definecolor{deltaBg}{RGB}{220,230,255} %

\newcommand{\method}{\textbf{Active-Zero}}
\newcommand{\scimp}[2]{#1\hspace{0.2em}\rlap{\scriptsize\textcolor{lightgreen}{#2}}}
\newcommand{\scdec}[2]{#1\hspace{0.2em}\rlap{\scriptsize\textcolor{gray!50}{#2}}}
\newcommand{\rowhighlight}{\rowcolor{deltaBg}}

\usepackage[preprint]{icml2026}

\usepackage{amsmath}
\usepackage{amssymb}
\usepackage{mathtools}
\usepackage{amsthm}
\usepackage{multirow}
\usepackage{arydshln}
\usepackage{graphicx}
\usepackage{amsfonts}
\usepackage{subcaption}
\usepackage{enumitem}

\usepackage[capitalize,noabbrev]{cleveref}

\theoremstyle{plain}

\theoremstyle{definition}

\theoremstyle{remark}

\usepackage[textsize=tiny]{todonotes}

\icmltitlerunning{Active Zero: Self-Evolving Vision-Language Models through Active Environment Exploration}

\begin{document}

\twocolumn[
  \icmltitle{Active Zero: Self-Evolving Vision-Language Models through Active Environment Exploration}
  \icmlsetsymbol{equal}{*}

  \begin{icmlauthorlist}
    \icmlauthor{Jinghan He}{1,2}
    \icmlauthor{Junfeng Fang}{3}
    \icmlauthor{Feng Xiong}{}
    \icmlauthor{Zijun Yao}{5}
    \icmlauthor{Fei Shen}{3}
    \icmlauthor{Haiyun Guo}{1,2} \\
    \icmlauthor{Jinqiao Wang}{1,2,4}
    \icmlauthor{Tat-Seng Chua}{3}
  \end{icmlauthorlist}

  \icmlaffiliation{1}{Foundation Model Research Center, Institute of Automation, Chinese Academy of Sciences}
  \icmlaffiliation{2}{School of Artificial Intelligence, University of Chinese Academy of Sciences}
  \icmlaffiliation{3}{National University of Singapore}
  \icmlaffiliation{4}{Wuhan AI Research}
  \icmlaffiliation{5}{Tsinghua University}

  \icmlcorrespondingauthor{}{hejinghan2022@ia.ac.cn. Code: \url{https://github.com/jinghan1he/Active-Zero}}

  \icmlkeywords{Machine Learning, ICML}

  \vskip 0.3in
]

\printAffiliationsAndNotice{} 

\input{sec/0_abstract}
\input{sec/1_introduction}
\input{sec/2_related_work}
\input{sec/3_method}
\input{sec/4_experiments}

\input{sec/5_conclusion}

\section*{Impact Statement}

This paper advances vision-language model reasoning through self-play and active exploration. The primary benefit is enabling more capable multimodal AI systems that can autonomously improve without extensive human annotation, potentially democratizing access to high-quality models.

We acknowledge potential risks: active exploration systems querying open-world repositories could inadvertently amplify biases or access inappropriate content, and self-improvement without human oversight raises alignment concerns. While our implementation uses curated datasets, future deployments to live web retrieval would require content filtering, bias mitigation, and safety guardrails. We encourage practitioners to implement appropriate oversight mechanisms when adopting active exploration paradigms.

\bibliography{main}
\bibliographystyle{icml2026}

\newpage
\appendix
\onecolumn

\input{sec/6_appendix}

\end{document}

%% file: sec/0_abstract.tex
\begin{abstract}
Self-play has enabled large language models to autonomously improve through self-generated challenges. However, existing self-play methods for vision-language models rely on passive interaction with static image collections, resulting in strong dependence on initial datasets and inefficient learning. Without the ability to actively seek visual data tailored to their evolving capabilities, agents waste computational effort on samples that are either trivial or beyond their current skill level. To address these limitations, we propose \method{}, a framework that shifts from passive interaction to active exploration of visual environments. \method{} employs three co-evolving agents: a Searcher that retrieves images from open-world repositories based on the model's capability frontier, a Questioner that synthesizes calibrated reasoning tasks, and a Solver refined through accuracy rewards. This closed loop enables self-scaffolding auto-curricula where the model autonomously constructs its learning trajectory. On Qwen2.5-VL-7B-Instruct across 12 benchmarks, \method{} achieves 53.97 average accuracy on reasoning tasks (5.7\% improvement) and 59.77 on general understanding (3.9\% improvement), consistently outperforming existing self-play baselines. These results highlight active exploration as a key ingredient for scalable and adaptive self-evolving vision-language systems.
\end{abstract}

%% file: sec/1_introduction.tex
\section{Introduction}

Self-play has emerged as a powerful paradigm for agent improvement, where systems bootstrap their capabilities by iteratively competing or collaborating with versions of themselves \cite{gao2025survey}. Originally demonstrated in game-playing domains \cite{silver2018general}, self-play enables agents to discover novel strategies and transcend their initial training data without human supervision. This principle of autonomous improvement has recently found new application in large language models (LLMs) \cite{chen2024self,wu2024self}.
Recent work has combined self-play with reinforcement learning with verifiable reward (RLVR) \cite{lambert2024tulu}, enabling LLMs to self-generate challenging tasks that push the boundaries of their own capabilities. Rather than relying solely on fixed datasets, these models autonomously create reasoning problems, attempt to solve them, and use the outcomes for self-improvement \cite{zhao2025absolute, huang2025r}.

\input{figs/intro}
 
Building on this success in language models, self-play is now extending into the multimodal domain, where vision-language models (VLMs) stand to benefit immensely from self-evolution given the vast visual corpora available on the internet \cite{radford2021learning}. Despite these advances, the transition from LLMs to VLMs introduces unique complexities: while LLMs operate in a self-contained symbolic space, VLMs must ground their reasoning in external visual environments \cite{wang2025perception}, requiring careful curation of visual experiences to enable effective self-evolution.

Current approaches address this by employing self-play over pre-collected image sets \cite{he2025visplay,thawakar2025evolmm}, synthesizing reasoning tasks from static visual galleries. While demonstrating initial promise, these frameworks \textbf{operate in a fundamentally passive mode}: they lack the ability to actively curate or explore visual data based on their evolving mastery. As illustrated in \cref{fig:intro} (a), this passivity creates two critical bottlenecks. First, it creates a \textbf{strong dependency on the initial image collection}—the quality and diversity of self-generated tasks are inherently bounded by the pre-selected visual data, regardless of the model's growing capabilities. Second, it leads to \textbf{learning inefficiency in open-world scenarios}. Without the ability to seek out images tailored to their current capability, agents expend computational effort on samples that are either trivial or beyond their skill level.

To overcome these bottlenecks, we propose a shift from passive interaction with static image galleries to active exploration of open-ended visual environments. As shown in \cref{fig:intro} (b), we realize this through \method{}, a tri-agent framework that closes the loop between data acquisition and reasoning enhancement via a three-stage training protocol. Our approach centers on the iterative optimization of three specialized agents, each addressing a distinct aspect of autonomous evolution. The cycle begins with the \textbf{Searcher}, which generates strategic queries to retrieve images from open-world repositories. By optimizing the Searcher to maximize the learning potential of retrieved data, we transform image selection from a static preprocessing step into a dynamic process driven by the model's evolving capabilities. The \textbf{Questioner} then formulates reasoning tasks calibrated to the frontier of the model's current abilities, ensuring the training signal remains neither trivial nor beyond reach. Finally, the \textbf{Solver} is refined via accuracy rewards to deepen its reasoning on these curated challenges. Through this iterative cycle, \method{} enables a self-scaffolding auto-curriculum where the model autonomously constructs its own learning trajectory, paving the way for more scalable and adaptive VLM reasoning.  

We evaluate \method{} across a comprehensive suite of benchmarks spanning 6 reasoning-intensive tasks and 6 general visual understanding. \method{} consistently outperforms self-play baselines, demonstrating that active environment exploration enables more effective auto-curriculum learning. On reasoning-intensive benchmarks, \method{} achieves 53.97 average accuracy on Qwen2.5-VL-7B-Instruct (a 5.7\% improvement), surpassing both VisPlay and EvolMM. These gains extend beyond narrow reasoning tasks--on general VLM benchmarks, \method{} achieves an average score of 59.77 (+3.9\%), compared to 57.51 for the base model. These results demonstrate that by dynamically selecting visual experiences aligned with its evolving capabilities, \method{} achieves simultaneous gains in both reasoning depth and general visual understanding. Our contributions are summarized as follows: 
\begin{itemize}[leftmargin=0.4cm]
    \item We fundamentally shift VLM self-play from passive interaction with static visual data to active exploration of visual environments, enabling agents to dynamically curate training data aligned with their capability frontier.
    \item We propose \method{}, a tri-agent framework that co-evolves a Searcher, Questioner, and Solver via iterative optimization, establishing a closed loop between open-world data discovery and reasoning improvement.
    \item Comprehensive evaluation demonstrates \method{} outperforms existing self-play baselines on both reasoning and general vision-language benchmarks.
\end{itemize}

%% file: figs/intro.tex
\begin{figure}[t]
	\centering
    \includegraphics[width=1\columnwidth]{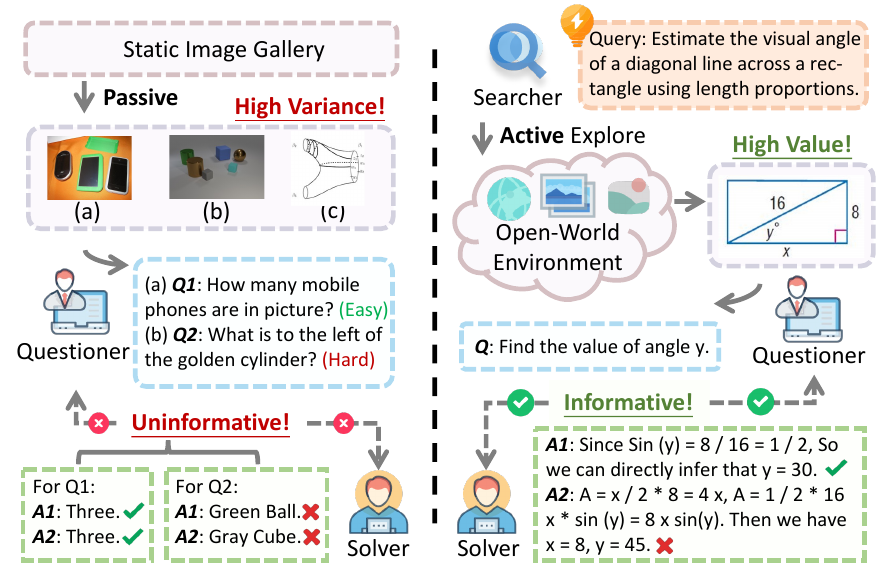}    
	\caption{
    \textbf{Comparison between Passive Approaches (left) and \method{} (right).} Unlike traditional methods limited by fixed data boundaries, \method{} actively explores open-world environments. It employs a Searcher to retrieve high-value images, forming a closed-loop auto-curriculum with the Questioner and Solver to scale VLM reasoning.
    }
    \label{fig:intro}
\end{figure} 

%% file: sec/2_related_work.tex
\section{Related Work}
\label{sec:related}

\subsection{Self-Evolving LLMs}
Recent work aims to remove the human-data bottleneck by enabling LLMs to self-generate training signals via self-play and verifiable rewards. Absolute Zero couples task proposal and solving in an RLVR loop grounded by executable environments, where tasks are programmatically validated against environment-derived ground truth \cite{zhao2025absolute}. R-Zero extends this to domains lacking explicit verifiers by replacing external verification with internal self-consistency: a Challenger generates questions targeting 50\% Solver agreement, while the Solver trains on difficulty-filtered majority-vote labels, creating an emergent curriculum that tracks the capability frontier \cite{huang2025r}. Beyond zero-data regimes, grounded self-play leverages external corpora to provide novelty and anchoring. SPELL enables long-context reasoning by cycling through questioner, responder, verifier roles, using semantic-equivalence verification and progressive length curricula for continual improvement \cite{yang2025spell}. SPICE generates question–answer pairs from documents, training a reasoner with variance-shaped rewards that track the capability frontier \cite{liu2025spice}. Complementary directions include self-rewarding frameworks that bootstrap RL from limited supervision \cite{fang2025serl}, game-theoretic alignment via iterative self-play \cite{wu2024self}, and curriculum-centric post-training that optimizes task selection for reasoning improvement \cite{chen2025self}.

\subsection{Active Data Curation}
Active data curation differs from conventional subset selection in that the learner acts to decide what data to obtain next, rather than sampling from a fixed pool. In vision-language settings, active learning variants select images or prompts using model-conditioned signals such as uncertainty or neighborhood structure to maximize improvement per labeled example \cite{safaei2025active,udandarao2025active}. Closer to open-world acquisition, target-driven dataset acquisition cast data collection as a planned process where an agent generates queries to retrieve task-suitable training data from external repositories \cite{berkane-etal-2025-llm}. A complementary direction studies adaptive curricula that reorder or schedule training data according to model-dependent difficulty, emphasizing that useful data changes as the model evolves \cite{zhang2025preference}. Our work aligns with the active side of this literature: rather than reweighting a static dataset, \method{} optimizes a Searcher to acquire challenging visual data from open-world repositories, then integrates acquisition with task synthesis and solving to form a closed-loop curriculum.

\subsection{Self-Evolving LVLMs}

Self-evolving LVLMs extend LLM self-play to visually grounded settings by generating tasks and rewards over images. VisPlay performs self-play on unlabeled images by co-training an image-conditioned Questioner and Reasoner with RL updates and difficulty-diversity shaping \cite{he2025visplay}. EvoLMM advances a similar Proposer--Solver loop but emphasizes continuous self-reward through consistency-based signals to drive improvement without external supervision \cite{thawakar2025evolmm}. Vision-zero reframes multimodal self-improvement as strategic, gamified self-play to better control task difficulty and mitigate plateaus \cite{wang2025vision}. Moving toward co-adaptation, C2-Evo explicitly co-evolves multimodal data complexity and model capability in a closed loop, targeting reasoning domains like multimodal math and geometry \cite{chen2025c2}. While these methods demonstrate the viability of self-play for VLMs, they operate over static image collections, treating the visual environment as fixed. In contrast, \method{} introduces active environment exploration: a learned Searcher dynamically retrieves images from open-world repositories based on the model's evolving capabilities, transforming the training environment into a controllable variable.

%% file: sec/3_method.tex
\input{figs/framework}

\section{Method}

\subsection{Preliminary}

Group Relative Policy Optimization (GRPO) \cite{shao2024deepseekmath} is an efficient reinforcement learning algorithm that optimizes models by eliminating the need for a separate value-function critic. For each input $q$, the old policy $\pi_{\theta_{old}}$ generates a group of $G$ independent responses $\{o_i\}_{i=1}^G$, where $o_{i}=\{o_{i,1},\cdots,o_{i,T_i}\}$ is a token sequence. Each response $o_i$ receives a reward $r_i$.
The policy $\pi_\theta$ is updated with a clipped objective at each token step $t$:
\begin{equation}
 \mathcal{L}_{i,t}^{clip}(\theta) = \min \left( r_{i,t}(\theta) \hat{A}_i, \text{clip}(r_{i,t}(\theta), \epsilon_{low}, \epsilon_{high}) \hat{A}_i \right),
\end{equation}
where the importance ratio $r_{i,t}(\theta)$ and the advantage $\hat{A}_i$ are:
\begin{equation}
r_{i,t}(\theta) = \frac{\pi_\theta(o_{i,t}|q, o_{i,<t})}{\pi_{\theta_{old}}(o_{i,t}|q, o_{i,<t})}, \space
\hat{A}_i = \frac{r_i - \text{mean}(\{r_k\}_{k=1}^G)}{\text{std}(\{r_k\}_{k=1}^G)}.
\end{equation}
In our framework, all three specialized agents are iteratively optimized using this mechanism to identify and bridge reasoning gaps.

\subsection{Framework Overview}
We introduce \method{}, a self-play framework that autonomizes the evolution of visual reasoning in VLMs through active environment exploration. Unlike passive approaches that rely on static image collections, \method{} actively explores open-world visual environments to construct its own training curriculum. The system comprises three specialized agents—the \textbf{Searcher} ($\mathcal{S}$), \textbf{Questioner} ($\mathcal{Q}$), and \textbf{Solver} ($\mathcal{V}$)—all initialized from a single base model but optimized for distinct roles. 
As illustrated in \cref{fig:framework}, the framework operates through a co-evolutionary cycle partitioned into three synergistic stages:
\begin{itemize}[leftmargin=0.4cm]
    \item \textbf{Active Environment Exploration (Stage 1):} $\mathcal{S}$ navigates an expansive visual environment $\mathcal{D}_{env}$ to retrieve images at the edge of the Solver's current capabilities—scenarios where the Solver is uncertain but not overwhelmed, ensuring high informational value.
    \item \textbf{Adaptive Task Synthesis (Stage 2):} $\mathcal{Q}$ transforms retrieved images into complex, multi-step reasoning questions $x$ calibrated to maximize the Solver's uncertainty, converting visual contexts into instructional curricula.
    \item \textbf{Reasoning Optimization (Stage 3):} $\mathcal{V}$ is optimized on the synthesized curriculum $\mathcal{D}_{train}$, through consensus-based reinforcement learning, bridging current reasoning gaps and raising the bar for subsequent iterations.
\end{itemize}

All three agents are optimized using GRPO, which employs a group-based advantage mechanism to filter noise and promote self-consistent reasoning trajectories. This unified RL backbone enables \method{} to autonomously scale reasoning complexity without exhaustive human annotation.

\subsection{Stage 1: Active Environment Exploration}

The Searcher $\mathcal{S}$ identifies visual scenarios that expose the Solver's current limitations. Formally, $\mathcal{S}$ is a policy $\pi_{\mathcal{S}}(q | P)$ that generates semantic retrieval queries $q$ conditioned on a domain-specific prompt $P$. Each query interacts with a retrieval function to fetch an image $I = \text{Ret}(q, \mathcal{D}_{env})$.

\textbf{Reward Formulation.} To prioritizes informative samples, we optimize $\mathcal{S}$ toward the frontier of the Solver's capabilities. Following \cite{huang2025r}, we define a challenge reward $R_{chal}$ based on the Solver’s predictive uncertainty. For a retrieved image $I$, we sample a probe question $x \sim \mathcal{Q}(\cdot|I)$ and execute $m$ independent reasoning passes through $\mathcal{V}$ to obtain an answer set $\{y_1, \dots, y_m\}$. The Solver’s empirical accuracy is then computed as:  
\begin{equation} 
\text{Acc}(\mathcal{V}, I, x) = \frac{1}{m} \sum_{j=1}^m \mathbb{I}[y_j = \hat{y}], 
\end{equation}
where $\hat{y}$ denotes the majority-voted consensus. The challenge reward peaks at maximum uncertainty:
\begin{equation}
\label{eq:reward_cha}
R_{chal}(q) = \mathbb{E}{x \sim \mathcal{Q}(\cdot|I)} \left[ 1 - 2 \left| \text{Acc}(\mathcal{V}, I, x) - \frac{1}{2} \right| \right].
\end{equation}
Maximizing $R_{chal}$ compels the Searcher to identify visual contexts at the Solver's capability frontier—tasks just beyond its deterministic grasp yet within its latent reasoning potential.

\textbf{Diversity Control.} To prevent the policy from converging on narrow query sets, we implement a dual-modality repetition penalty $\mathcal{P}_{rep}$. For each batch $\mathbf{Q} = \{q_i\}_{i=1}^n$ with retrieved images $\mathbf{I} = \{I_i\}_{i=1}^n$, we quantify pairwise similarity in both text and visual spaces.
\begin{equation}
\begin{aligned}
d_{ij}^{txt} &= 1 - \text{BLEU}(q_i, q_j), \\
d_{ij}^{vis} &= 1 - \cos(\phi(I_i), \phi(I_j)),
\end{aligned}
\end{equation}
where $\phi(\cdot)$ represents a frozen image encoder. We cluster a batch of samples into $\mathcal{C}^{txt}$ and $\mathcal{C}^{vis}$ based on sensitivity thresholds $\tau_{txt}$ and $\tau_{vis}$. The penalty for query-image pair $(q_i, I_i)$ is then proportional to neighborhood density:  
\begin{equation}
\label{eq:p_rep}
\mathcal{P}_{rep}(q_i, I_i) = \frac{|C^{txt}(q_i)|}{n} + \frac{|C^{vis}(I_i)|}{n}.
\end{equation}
The final reward is $R_{\mathcal{S}} = \max(0, R_{chal} - \mathcal{P}_{rep})$, encouraging the agent to explore diverse coordinates across the visual-semantic manifold.

\textbf{Domain-Conditioned Exploration.} To maintain a balanced curriculum across heterogeneous tasks, we partition the search space into $N$ semantic domains: $\mathcal{C} = \{c_{1}, c_{2}, \cdots, c_{N}\}$. Each domain is governed by a specialized prompt template $P_c$ that constrains $\pi_{\mathcal{S}}$ to relevant visual-logical properties. To prevent the model from ignoring challenging domains in favor of easier ones with higher absolute rewards, we utilize the group-relative nature of GRPO. For a group of $G$ queries within domain $c$, the advantage $\hat{A}$ is normalized strictly against domain-specific peers:
\begin{equation}
\hat{A}(q_{i,c}) = \frac{R_{\mathcal{S}}(q_{i,c}) - \mu_c}{\sigma_c}.
\end{equation}
This ensures that the Searcher seeks the most informative data within every domain, sustaining a robust auto-curriculum across the multi-modal spectrum.

\subsection{Stage 2: Adaptive Task Synthesis}

Following optimization of the search policy, we deploy $\pi_{\mathcal{S}}^*$ to curate an active visual dataset $\mathcal{D}_{active} = \{I_j\}_{j=1}^M$. The Questioner $\pi_{\mathcal{Q}}(x | I, T_{cot})$ synthesizes complex reasoning tasks $x$ from these images. We employ a Chain-of-Thought (CoT) template $T_{cot}$ to guide the Questioner to deconstruct visual-logical constraints prior to question formulation.

\textbf{Reward Formulation.} Similar to the Searcher, the Questioner aims to maximize the instructional utility of the generated questions by targeting the Solver's current weaknesses. The reward $R_{\mathcal{Q}}$ for question $x_i$ generated from image $I$ is:
\begin{equation}
R_{\mathcal{Q}}(x_i) = \max(0, R_{chal}(x_i | I) - \mathcal{P}_{rep}(x_i)),
\end{equation}
where $R_{chal}(x_i | I)$ is the challenge reward from \cref{eq:reward_cha} for question $x_i$, and $\mathcal{P}_{rep}$ is a linguistic repetition penalty preventing collapse into repetitive question structures. 

By maximizing $R_{chal}$, the Questioner generates tasks that avoid both trivial consensus and complete randomness, creating a curriculum that bridges the Solver's current state and higher-order reasoning capabilities.  

\subsection{Stage 3: Reasoning Optimization}

\textbf{Training Set Construction.} In the final stage, we deploy the optimized policy $\pi_{\mathcal{Q}}^*$ over $\mathcal{D}_{active}$ to construct an initial training pool $\mathcal{D}_{train} = \{(I_j, x_j)\}_{j=1}^M$. For each task, we compute a silver-standard consensus answer $\hat{a}$ by majority vote over $K$ sampled reasoning trajectories from $\pi_{\mathcal{V}}$ . To focus on informative samples, we filter the pool to retain tasks where empirical accuracy $\text{Acc}(\mathcal{V}, I, x)$ falls within difficulty window $(\tau_{low}, \tau_{high})$, yielding curated subset $\mathcal{D}_{train}^*$. This ensures the Solver optimize trajectories that are neither trivially solved nor beyond its current capacity.

\textbf{Reward Formulation.} The Solver is optimized via GRPO to maximize the accuracy reward signal against pseudo-labels. For each task $(I, x) \in \mathcal{D}_{train}^*$, we sample a group of $G$ independent reasoning trajectories $\{\tau_1, \dots, \tau_G\}$, where each $\tau_i$ consists of a CoT path and final answer $a_i$. The reward for a trajectory $\tau_i$ is defined as:
\begin{equation}
R_{\mathcal{V}}(\tau_i) =
\begin{cases}
1, & \text{if } a_i = \hat{a}, \\
0, & \text{otherwise}.
\end{cases}
\end{equation}
This process ultimately enhances the Solver’s ability to navigate the complex visual reasoning challenges synthesized by the preceding agents.

%% file: figs/framework.tex
\begin{figure*}[t]
% \vspace{-20pt}
	\centering
    \includegraphics[width=\textwidth]{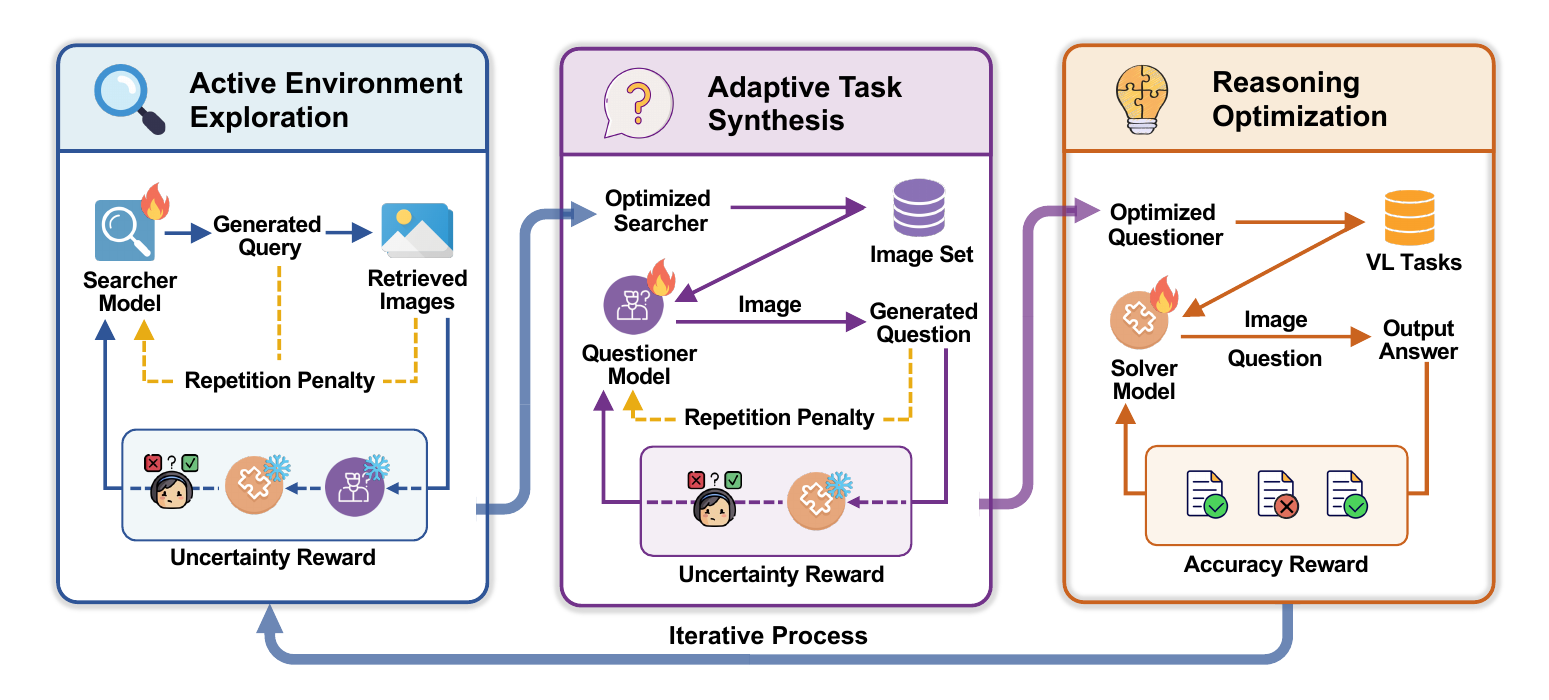}
	\caption{
    \textbf{Overview of the \method{} framework.} The system facilitates an iterative, three-stage self-play cycle to enhance VLM reasoning: \textbf{(Stage 1)} The \textbf{Searcher} ($\mathcal{S}$) is optimized to retrieve informative images from the environment by balancing a challenge reward (targeting the Solver's uncertainty) and a repetition penalty. \textbf{(Stage 2)} The \textbf{Questioner} ($\mathcal{Q}$) synthesizes complex, multi-step questions from the curated images to maximize instructional utility. \textbf{(Stage 3)} The \textbf{Solver} ($\mathcal{V}$) is trained on the generated tasks to improve reasoning accuracy and consensus via reinforcement learning. The entire process is iterative, allowing the agents to co-evolve as the Solver’s capability frontier shifts.
    }
	\label{fig:framework}
\end{figure*} 

%% file: sec/4_experiments.tex
\section{Experiments}

\input{tables/main_table1}
\input{tables/main_table2}

\subsection{Implementation Details}

\textbf{Visual Environment Setup.} To facilitate efficient and reproducible training, we construct a diverse visual environment $\mathcal{D}_{env}$ by aggregating the image from The Cauldron \cite{laurençon2024matters}, a comprehensive collection of 50 vision-language datasets, and three geometry-focused datasets: Geo3K~\cite{lu2021inter}, UniGeo~\cite{chen2022unigeo}, and GeoQA+~\cite{cao-xiao-2022-augmented}. The resulting $\mathcal{D}_{env}$ contains 1.6M images, providing a controlled testbed for analyzing the Searcher’s behavior and while ensuring reproducibility without the temporal stochasticity and latency of real-time web APIs.

\textbf{Retrieval Infrastructure.} The Searcher $\mathcal{S}$ interfaces with the environment via a vector retrieval system. We use SigLIP-2 \cite{tschannen2025siglip} to generate visual embeddings for all images in $\mathcal{D}_{env}$, indexed using FAISS \cite{douze2025faiss} for sub-millisecond search. When the Searcher generates query $q$, the retrieval function $\text{Ret}(q, \mathcal{D}_{env})$ computes the cosine similarity between the query embedding and the visual index to return the top-k candidate.

\textbf{System Scalability.} While our implementation operates on this pre-indexed local structure for computational efficiency, the Searcher's query-based interface is backend-agnostic. This architecture enables seamless transition to real-time web exploration or specialized domain-specific databases by swapping the retrieval backend. The million-level local repository provides sufficient diversity to challenge the Searcher’s discovery capabilities while maintaining stability required for iterative optimization.

\input{figs/dynamics}
\input{tables/ablation}

\textbf{Optimization Hyperparameters.} All three agents are optimized using GRPO \cite{shao2024deepseekmath} with a group size of $G=8$. Clustering thresholds are $\tau_{txt} = 0.5$ and $\tau_{vis} = 0.1$. To construct $\mathcal{D}_{active}$, we sample 6000 queries from $\pi_{\mathcal{S}}$ and retrieve the top $5$ images per query from $\mathcal{D}_{env}$, followed by a deduplication. For each image in $\mathcal{D}_{active}$, one question is generated from $\pi_{\mathcal{Q}}$ to form $\mathcal{D}_{train}$. The difficulty filtering thresholds are $\tau_{low}=0.3$ and $\tau_{high}=0.8$. Training is conducted on 16 NVIDIA H20 GPUs using the EasyR1 library \cite{zheng2025easyr1}. Additional Details are in \cref{sec:setup}.

\textbf{Evaluation Protocol.}  We use a consistent CoT prompt for all models and report greedy decoding results. Predefined format-matching rules are used to extract final answers from model responses. To avoid limitations of rigid string matching, we use DeepSeek-V3.2 \cite{liu2025deepseek} as an automated evaluator. The LLM receives the question, ground truth, and extracted answer to judge semantic and logical consistency, ensuring that correct reasoning is not penalized for formatting variations. We provide the benchmark details in \cref{sec:setup} and evaluation prompts in \cref{sec:prompt}.

\subsection{Main Results}

We evaluate \method{} across a comprehensive suite of benchmarks. As shown in \cref{tab:main_table1,tab:main_table2}, we compare our framework against the Base Model and recent self-play baselines, including VisionZero~\cite{wang2025vision}, VisPlay~\cite{he2025visplay}, and EvolMM~\cite{thawakar2025evolmm}.

\textbf{Reasoning-Intensive Benchmarks.} As shown in \Cref{tab:main_table1}, \method{} delivers the strongest overall results at both model scales. It raises the average score on Qwen2.5-VL-3B-Instruct from 41.34 to 45.33 (+3.99) and on Qwen2.5-VL-7B-Instruct from 51.05 to 53.97 (+2.92). Improvements are consistent across all six benchmarks. For the 7B model, the largest gains come from LogicVista (48.21 vs.\ 43.08) and WeMath (69.60 vs.\ 64.48). Notably, \method{} also outperforms all prior self-play methods on the 7B setting, reaching a 53.97 average versus 52.02 for VisPlay and 51.39 for EvolMM, highlighting the benefit of uncertainty-driven retrieval and task synthesis over existing curricula.

\input{figs/question_dist}

\textbf{General Visual Understanding.} In \cref{tab:main_table2}, \method{} also improves general VLM performance, reaching the best average on Qwen2.5-VL-7B-Instruct (59.77 vs.\ 57.51 for the base model). Improvements are consistent across VisNum, RealWorldQA, MMStar, MMMU, and MMMU-Pro (e.g., 47.94 on VisNum and 67.58 on RealWorldQA), while Hallusion remains comparable but slightly lower than the base model (68.03 vs.\ 68.87). On Qwen2.5-VL-3B-Instruct, \method{} similarly increases the average from 49.32 to 52.55, with particularly strong performance on VisNum (41.51) and MMMU-Pro (46.70).

\subsection{Effectiveness of Active Environment Exploration}

We study the contribution of the Searcher $\mathcal{S}$ using the 3B setting as a controlled baseline (\cref{tab:ablation}).

\textbf{Searcher vs.\ Random Sampling.} Replacing \method{} with random image retrieval reduces the reasoning average from 45.33 to 43.88. This drop (1.45 points) indicates that actively selecting informative samples is substantially more effective than uniform sampling from $\mathcal{D}_{env}$ for improving reasoning. A similar degradation is observed in general performance (52.55 → 51.86), suggesting that informed retrieval benefits both reasoning-specific and broader visual understanding objectives.

\textbf{Effect of RL Optimization.} Disabling RL optimization in the Searcher leads to a consistent degradation, with reasoning dropping to 44.75 and general performance to 51.03. Compared to the full method (45.33 / 52.55), this suggests that RL-based selection better matches the Solver’s learning needs than heuristic or unoptimized retrieval. Notably, the random variant attains a higher general average (51.86) than the unoptimized Searcher (51.03), implying that without RL, the Searcher may over-focus on samples that benefit reasoning at the expense of broad coverage.

\input{figs/image_dist}

\subsection{Ablation Study}

We further validate two key design choices in the lower section of \cref{tab:ablation}.

\textbf{Domain-Conditioned Exploration.} Removing domain conditioning decreases both reasoning (45.33$\rightarrow$44.93) and general performance (52.55$\rightarrow$50.73). The larger drop on general benchmarks suggests that domain-aware partitioning helps maintain a balanced curriculum across heterogeneous domains, preventing skewed exploration.

\textbf{Visual Redundancy Control.} We ablate the visual repetition penalty $\mathcal{P}_{rep}$ while keeping other components unchanged. Removing it reduces the reasoning average to 44.89 and the general average to 51.62. This supports the role of visual diversity control in preventing the Searcher from repeatedly sampling near-duplicate images, thereby improving coverage of the visual-semantic space needed for robust reasoning.

\subsection{Training Dynamics}

To better understand the self-play process, we visualize the co-evolution of the Solver, the Searcher, and the reward signals during training in \cref{fig:dynamics}.

\textbf{Solver evolution and the benefit of active sampling.} In \cref{fig:dy_v}, \method{} consistently achieves higher validation accuracy than the random-retrieval baseline, with the gap becoming more apparent in later training steps. This trend suggests that uncertainty-driven sample selection provides a stronger curriculum than uniform sampling, especially as the Solver becomes capable of benefiting from harder but still learnable examples.

\textbf{Searcher learning signal across phases.} \Cref{fig:dy_s} shows that the uncertainty reward rises across both training phases (v1 and v2), suggesting the Searcher increasingly retrieves examples informative for the current Solver. Meanwhile, \cref{fig:dy_sd} shows the diversity penalty drops early and then stabilizes, indicating the Searcher avoids collapsing onto near-duplicate high-uncertainty images. Overall, it maintains broad coverage while improving the reward signal, balancing informativeness and diversity.

\subsection{Discussion}

\textbf{Active exploration yields healthier difficulty distribution.} To diagnose how exploration reshapes the curriculum, we analyze the consensus distribution of generated questions in \cref{fig:ques_dist}. The consensus score measures agreement among multiple Solver rollouts. Low consensus indicates unstable or overly difficult questions, high consensus reflects reliably solved questions, and intermediate consensus corresponds to moderately challenging samples near the Solver's capability frontier.
Compared to Random Sampling, \method{} shifts mass away from the lowest-consensus bin and increases mid-to-high consensus questions across versions (v1--v3). Random Sampling concentrates heavily in the minimum-consensus bin, indicating that uniform retrieval often produces noisy or weakly grounded questions with inconsistent training signals. This validates that active retrieval better supports coherent, learnable questions.

\textbf{Active exploration yields naturally diverse curricula.} As illustrated in \cref{fig:image_dist}, \method{} Searcher distributes retrievals across diverse task categories: Diagrams/Geometry (43\%), Text/OCR (12.7-14.3\%), Visual Reasoning (9.4-11.9\%), and multiple others, demonstrating that active retrieval naturally discovers varied visual reasoning challenges. Domain-conditioned exploration (left) further stabilizes this balance, while its removal (right) causes gradual concentration toward easier categories, validating the robustness of the active exploration framework.

\textbf{Query evolution reflects adaptive curriculum construction.}
We analyze how the Searcher's retrieval strategy evolves by tracking query token frequency changes across training versions (\cref{fig:word_cloud}). From v1 to v2, the Searcher progressively adopts more structural and process-oriented vocabulary, with increased use of terms like \emph{field}, \emph{direction}, \emph{diagram}, \emph{forces}, \emph{folded}, \emph{magnetic}. This shift indicates movement from generic visual descriptors toward images encoding spatial relationships, physical properties, and multi-component structures that demand integrative reasoning. The emergence of technical and relational terminology suggests the Searcher adapts queries in response to the Solver's learning progress, increasingly targeting images with richer structural information. This qualitative evolution aligns with the consensus analysis in \cref{fig:ques_dist}, where active exploration reduces unstable samples and concentrates toward more learnable questions at the Solver's capability frontier, demonstrating that the framework constructs increasingly sophisticated curricula as training progresses.

\input{figs/word_cloud}

%% file: tables/main_table1.tex
\begin{table*}[t!]
\centering
\caption{Comparison of \method{} against state-of-the-art baselines on reasoning-intensive benchmarks.}
\label{tab:main_table1}
\small
\resizebox{\textwidth}{!}{
\begin{tabular}{p{3cm}p{1.6cm}p{1.6cm}p{1.6cm}p{1.6cm}p{1.6cm}p{1.6cm}p{1.6cm}p{1.6cm}}
\toprule[1.5pt]
\textbf{Method} & \textbf{MathVista} & \textbf{MathVision} & \textbf{WeMath} & \textbf{MathVerse} & \textbf{LogicVista} & \textbf{DynaMath} & \textbf{Averge}\\
\midrule
\midrule
\multicolumn{9}{@{}l}{\textit{Qwen2.5-VL-3B-Instruct}} \\
\cdashline{1-8}[2pt/2pt]
\rule{0pt}{10pt}%
 Base Model &
57.80 & 22.27 & 53.51 & 30.84 & 37.28 & 46.33 & 41.34 \\
\cdashline{1-8}[2pt/2pt]
\rule{0pt}{10pt}%
 VisPlay &
\scimp{58.10}{+0.30} &
\scimp{22.43}{+0.16} &
\scimp{58.16}{+4.65} &
\scimp{34.58}{+3.74} &
\scimp{39.29}{+2.01} &
\scimp{50.02}{+3.69} &
\scimp{43.76}{+2.42} \\
\rowhighlight
 \method{} &
\scimp{\textbf{60.20}}{+2.40} &
\scimp{\textbf{23.62}}{+1.35} &
\scimp{\textbf{61.21}}{+7.70} &
\scimp{\textbf{35.09}}{+4.25} &
\scimp{\textbf{40.62}}{+3.34} &
\scimp{\textbf{51.24}}{+4.91} &
\scimp{\textbf{45.33}}{+3.99} \\
\midrule
\multicolumn{9}{@{}l}{\textit{Qwen2.5-VL-7B-Instruct}} \\
\cdashline{1-8}[2pt/2pt]
\rule{0pt}{10pt}%
 Base Model &
69.40 & 25.95 & 64.48 & 45.11 & 43.08 & 58.26 & 51.05 \\
\cdashline{1-8}[2pt/2pt]
\rule{0pt}{10pt}%
 VisionZero-CLEVR &
\scimp{70.20}{+0.80} &
\scdec{24.67}{-1.28} &
\scimp{66.90}{+2.42} &
\scdec{45.05}{-0.06} &
\scdec{41.29}{-1.79} &
\scimp{58.70}{+0.44} &
\scimp{51.14}{+0.09} \\
 VisionZero-Chart &
\scdec{68.20}{-1.20} &
\scdec{25.16}{-0.79} &
\scimp{64.66}{+0.18} & 
\scimp{46.57}{+1.46} &
\scimp{46.21}{+3.13} &
\scimp{58.40}{+0.14} &
\scimp{51.53}{+0.48} \\
 VisionZero-RealWorld &
\scdec{69.20}{-0.20} &
\scimp{26.12}{+0.17} &
\scimp{65.40}{+0.92} &
\scimp{45.62}{+0.51} &
\scimp{\textbf{48.44}}{+5.36} &
\scimp{58.62}{+0.36} &
\scimp{52.23}{+1.18} \\
 VisPlay &
\scimp{69.40}{+0.00} &
\scimp{\textbf{27.04}}{+1.09} &
\scimp{67.99}{+3.51} &
\scdec{43.59}{-1.52} &
\scimp{45.09}{+2.01} &
\scimp{58.98}{+0.72} &
\scimp{52.02}{+0.97} \\
 EvoLMM &
\scimp{69.80}{+0.40} &
\scdec{24.11}{-1.84} &
\scimp{64.89}{+0.41} &
\scdec{44.80}{-0.31} &
\scimp{46.65}{+3.57} &
\scdec{58.06}{-0.20} &
\scimp{51.39}{+0.34} \\
 \rowhighlight
 \method{} &
\scimp{\textbf{72.60}}{+3.20} &
\scimp{26.18}{+0.23} &
\scimp{\textbf{69.60}}{+5.12} &
\scimp{\textbf{48.16}}{+3.05} &
\scimp{48.21}{+5.13} &
\scimp{\textbf{59.04}}{+0.78} &
\scimp{\textbf{53.97}}{+2.92} \\
\bottomrule[1.5pt]
\end{tabular}}
\end{table*}

%% file: tables/main_table2.tex
\begin{table*}[t!]
\centering
\caption{Comparison of \method{} against state-of-the-art baselines on general visual understanding and knowledge-based benchmarks.}
\label{tab:main_table2}
\small
\resizebox{\textwidth}{!}{
\begin{tabular}{p{3cm}p{1.6cm}p{1.6cm}p{1.6cm}p{1.6cm}p{1.6cm}p{1.6cm}p{1.6cm}p{1.6cm}}
\toprule[1.5pt]
\textbf{Method} & \textbf{VisNum} & \textbf{RealWorld} & \textbf{MMStar} & \textbf{MMMU} & \textbf{MMMU\textunderscore{}Pro} & \textbf{Hallusion} & \textbf{Averge}\\
\midrule
\midrule
\multicolumn{9}{@{}l}{\textit{Qwen2.5-VL-3B-Instruct}} \\
\cdashline{1-8}[2pt/2pt]
\rule{0pt}{10pt}%
 Base Model               & 34.71 & \textbf{61.70} & 52.80 & 44.69 & 41.99 & 60.04 & 49.32 \\
\cdashline{1-8}[2pt/2pt]
\rule{0pt}{10pt}%
 VisPlay & 
\scimp{37.32}{+2.61} & 
\scdec{59.48}{-2.22} & 
\scdec{51.60}{-1.20} & 
\scimp{44.80}{+0.11} & 
\scimp{42.43}{+0.44} & 
\scimp{64.88}{+4.84} & 
\scimp{50.09}{+0.77} \\
\rowhighlight
\method{} & 
\scimp{\textbf{41.51}}{+6.80} & 
\scdec{60.00}{-1.70} & 
\scimp{\textbf{53.80}}{+1.00} & 
\scimp{\textbf{49.16}}{+4.47} & 
\scimp{\textbf{46.70}}{+4.71} & 
\scimp{\textbf{64.14}}{+4.10} & 
\scimp{\textbf{52.55}}{+3.23} \\

\midrule
\multicolumn{9}{@{}l}{\textit{Qwen2.5-VL-7B-Instruct}} \\
\cdashline{1-8}[2pt/2pt]
\rule{0pt}{10pt}%
 Base Model               & 44.85 & 63.66 & 61.53 & 53.52 & 52.61 & 68.87 & 57.51 \\
\cdashline{1-8}[2pt/2pt]
\rule{0pt}{10pt}%
VisionZero-CLEVR & 
\scdec{43.13}{-1.72} & 
\scimp{66.41}{+2.75} & 
\scimp{62.07}{+0.54} & 
\scimp{\textbf{56.98}}{+3.46} & 
\scdec{52.17}{-0.44} & 
\scimp{69.51}{+0.64} & 
\scimp{58.38}{+0.87} \\

VisionZero-Chart & 
\scdec{44.54}{-0.31} & 
\scimp{64.05}{+0.39} & 
\scimp{62.27}{+0.74} & 
\scdec{53.07}{-0.45} & 
\scdec{51.67}{-0.94} & 
\scimp{69.09}{+0.22} & 
\scdec{57.45}{-0.06} \\

VisionZero-RealWorld & 
\scdec{44.17}{-0.68} & 
\scimp{65.36}{+1.70} & 
\scimp{62.27}{+0.74} & 
\scdec{53.30}{-0.22} & 
\scdec{52.29}{-0.32} & 
\scdec{68.56}{-0.31} & 
\scimp{57.66}{+0.15} \\

VisPlay & 
\scimp{45.22}{+0.37} & 
\scdec{63.27}{-0.39} & 
\scimp{62.93}{+1.40} & 
\scdec{52.51}{-1.01} & 
\scdec{51.98}{-0.63} & 
\scdec{68.56}{-0.31} & 
\scdec{57.41}{-0.10} \\

EvoLMM & 
\scimp{45.01}{+0.16} & 
\scdec{61.44}{-2.22} & 
\scimp{61.53}{+0.00} & 
\scimp{55.31}{+1.79} & 
\scdec{51.60}{-1.01} & 
\scimp{\textbf{69.72}}{+0.85} & 
\scdec{57.44}{-0.07} \\
\rowhighlight
\method{} & 
\scimp{\textbf{47.94}}{+3.09} & 
\scimp{\textbf{67.58}}{+3.92} & 
\scimp{\textbf{64.00}}{+2.47} & 
\scimp{56.42}{+2.90} & 
\scimp{\textbf{54.62}}{+2.01} & 
\scdec{68.03}{-0.84} & 
\scimp{\textbf{59.77}}{+2.26} \\

\bottomrule[1.5pt]
\end{tabular}}
\end{table*}

%% file: figs/dynamics.tex
\begin{figure*}[t]
	\centering
    \begin{subfigure}[b]{0.31\linewidth}
        \centering
		\includegraphics[width=\textwidth]{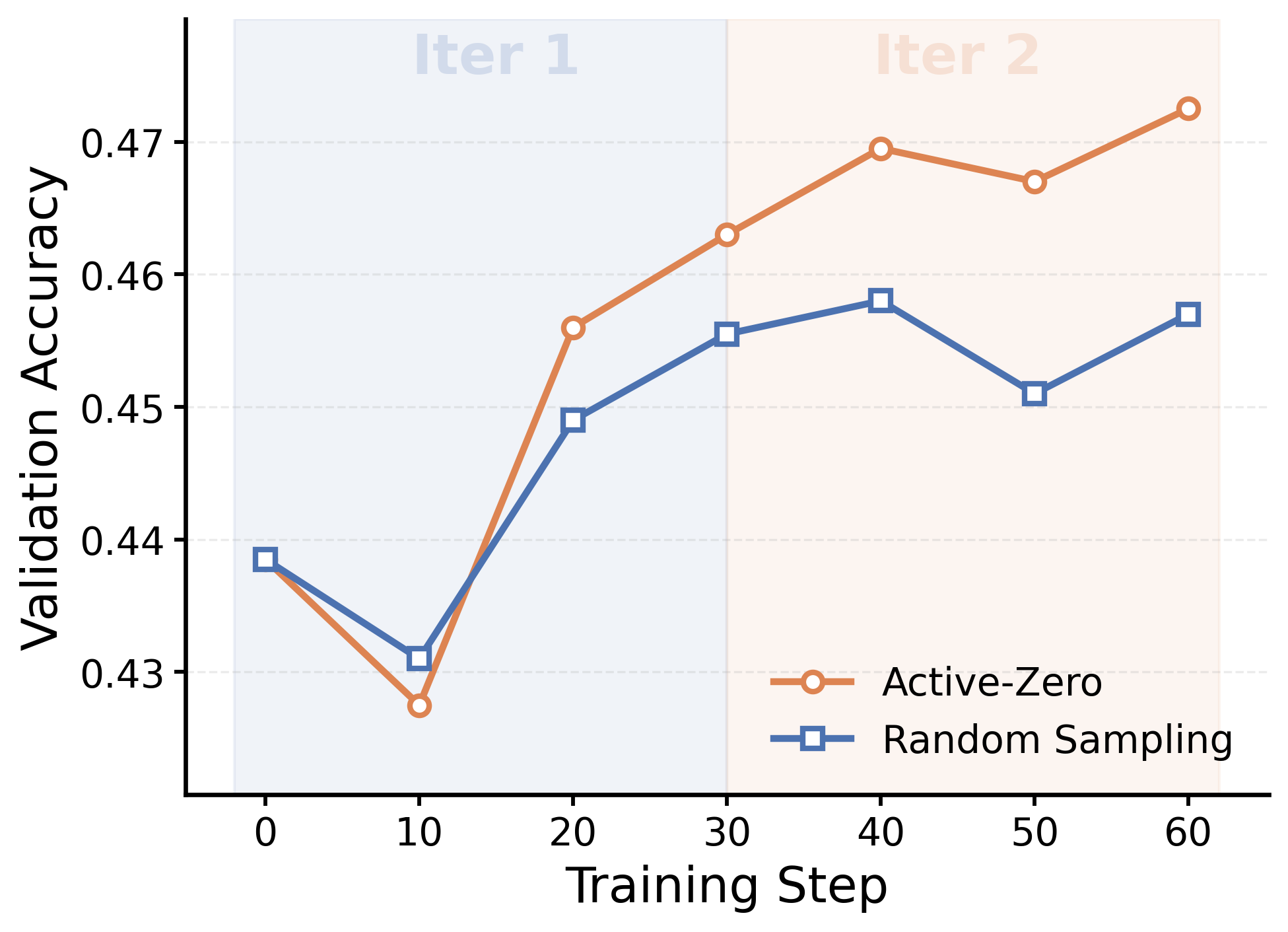}
        \caption{Solver Accuracy}
	    \label{fig:dy_v}
    \end{subfigure}
    \hfill
	\begin{subfigure}[b]{0.31\linewidth}
        \centering
		\includegraphics[width=\textwidth]{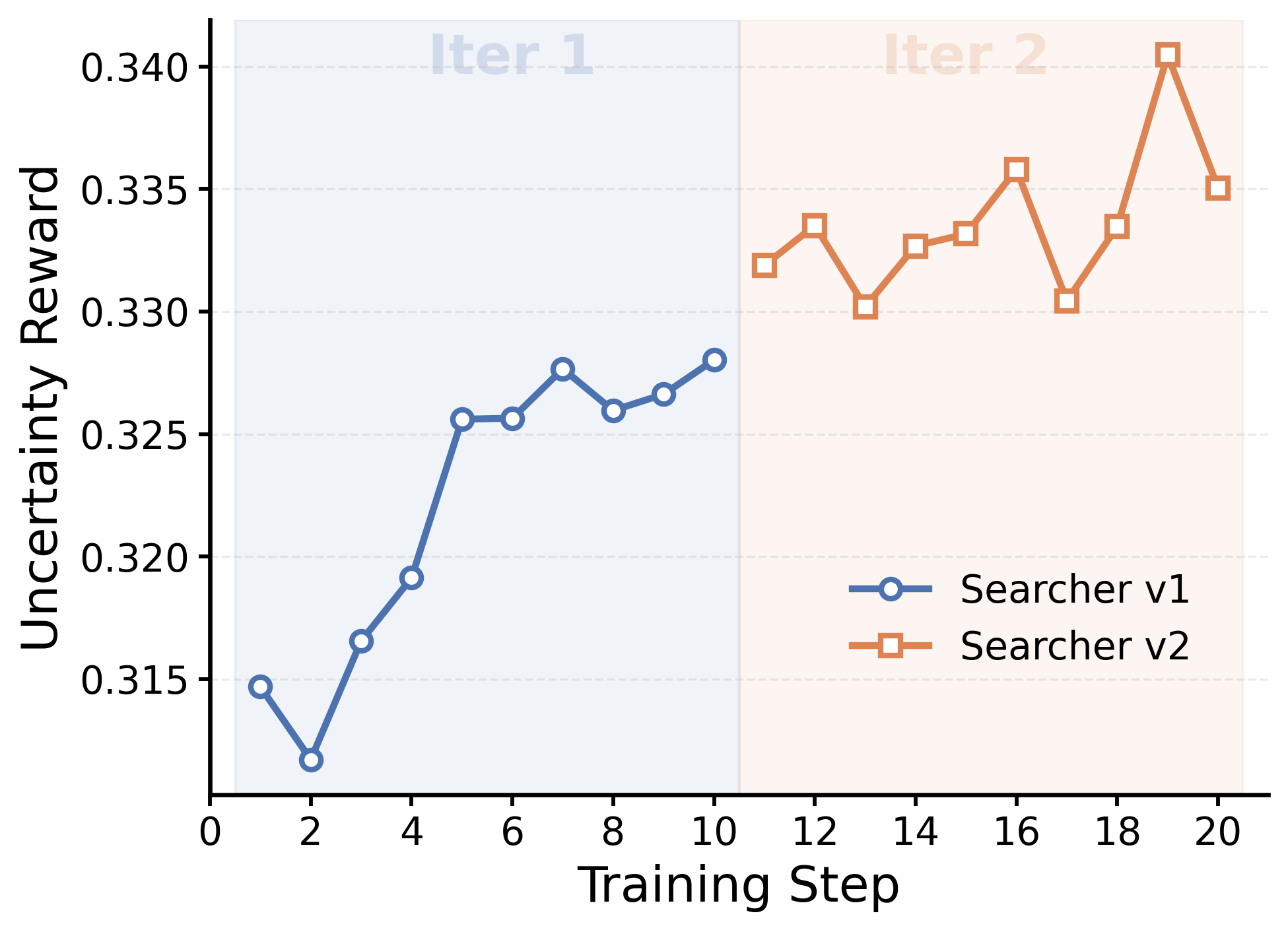}
        \caption{Searcher Reward}
	    \label{fig:dy_s}
    \end{subfigure}
    \hfill
    \begin{subfigure}[b]{0.31\linewidth}
        \centering
		\includegraphics[width=\textwidth]{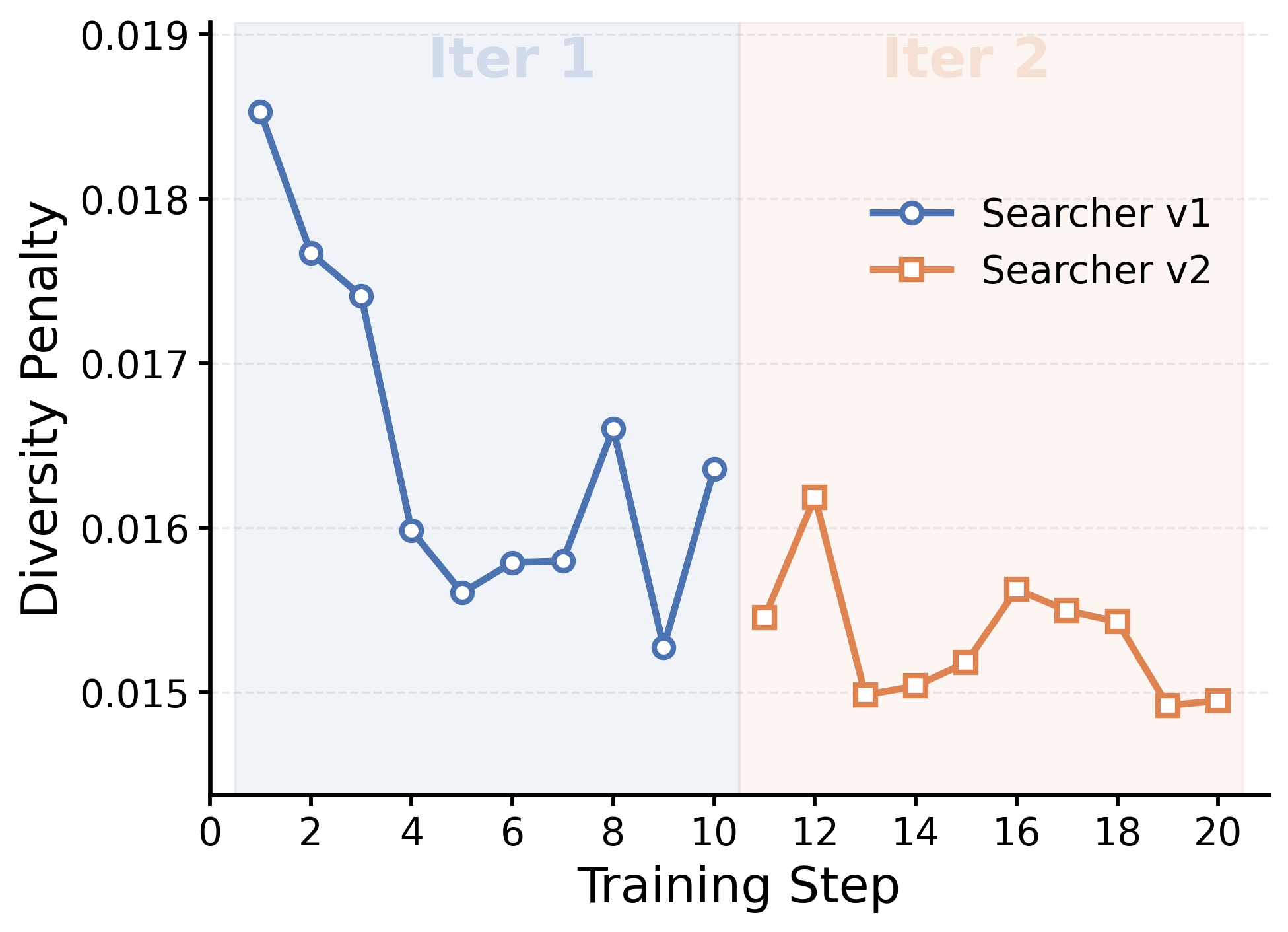}
        \caption{Searcher Diversity}
	    \label{fig:dy_sd}
    \end{subfigure}
	\caption{
    \textbf{Co-evolution of Solver and Searcher.} Active-Zero shows superior accuracy scaling in (a), driven by the Searcher's ability to maximize uncertainty (b) while maintaining high sample diversity (c) through RL-based optimization.
    }
    \label{fig:dynamics}
\end{figure*} 

%% file: tables/ablation.tex
\begin{table}[t]
\centering
\caption{Ablation study on \method{} components.}
\label{tab:ablation}
\resizebox{\columnwidth}{!}{
\begin{tabular}{
l
>{\centering\arraybackslash}p{0.3\columnwidth}
>{\centering\arraybackslash}p{0.3\columnwidth}
}
\toprule[1.5pt]
\textbf{Method} & \textbf{Reasoning Avg.} & \textbf{General Avg} \\
\midrule
\midrule
\method{} & \textbf{45.33} & \textbf{52.55} \\
\midrule
\textit{Searcher Effectiveness}     & & \\
\quad $\vdash$ Base Searcher      & \scdec{44.75}{-0.58} & \scdec{51.03}{-1.52} \\
\quad $\vdash$ Random Sampling    & \scdec{43.88}{-1.45} & \scdec{51.86}{-0.69} \\
\midrule
\textit{Ablations} & & \\
\quad $\vdash$ w/o Domain-Conditioned      & \scdec{44.93}{-0.40} & \scdec{50.73}{-1.82} \\
\quad $\vdash$ w/o Visual Penalty          & \scdec{44.89}{-0.44} & \scdec{51.62}{-0.93} \\
\bottomrule[1.5pt]
\end{tabular}
}
\end{table}

%% file: figs/question_dist.tex
\begin{figure}[t]
	\centering
    \includegraphics[width=\columnwidth]{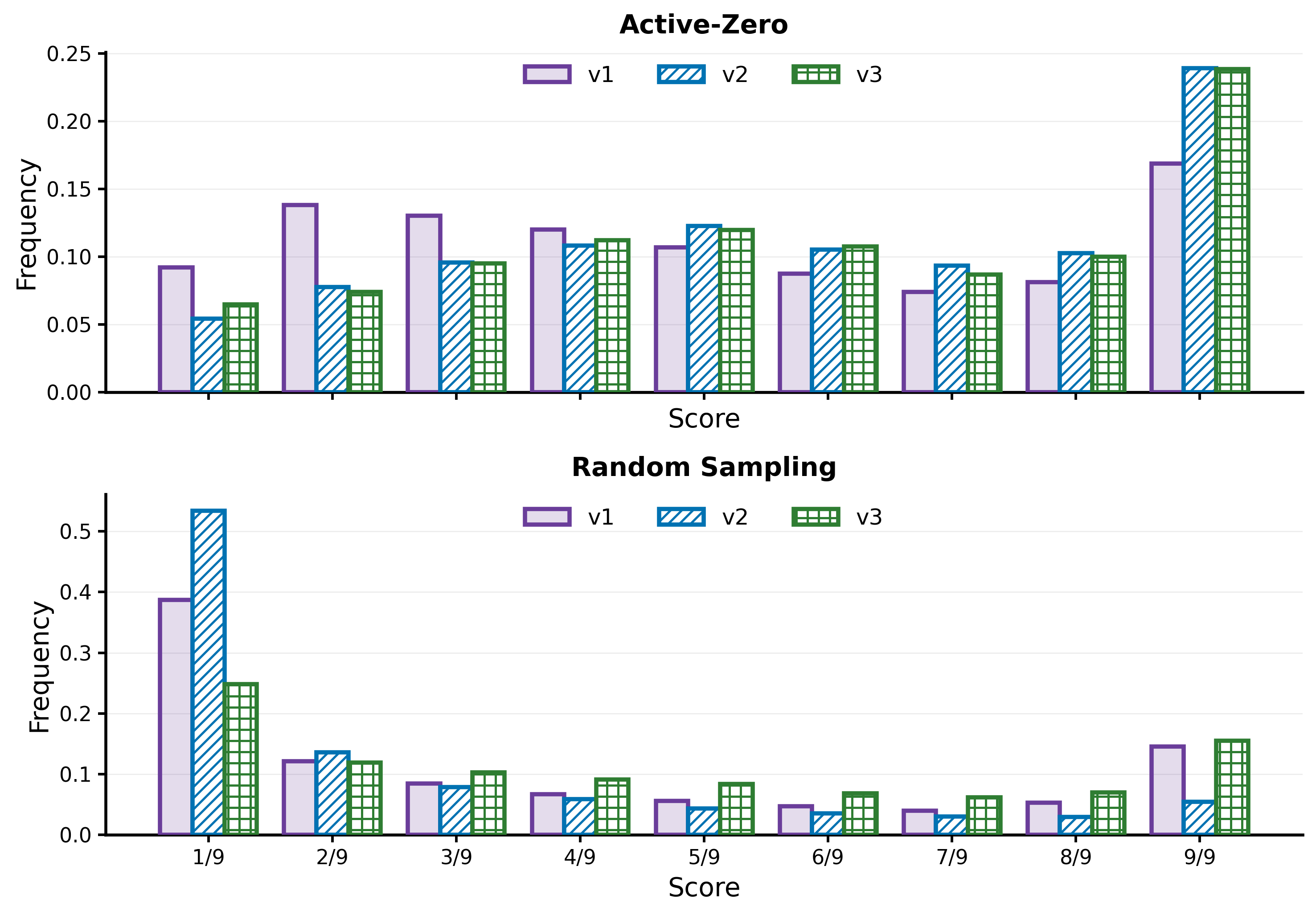}   
	\caption{
    \textbf{Consensus-score distribution of generated VL tasks} under \method{} and Random Sampling across training versions (v1–v3). The consensus score measures agreement among multiple Solver rollouts on the same question.
    }
    \label{fig:ques_dist}
\end{figure}

%% file: figs/image_dist.tex
\begin{figure*}[t]
	\centering
    \includegraphics[width=\linewidth]{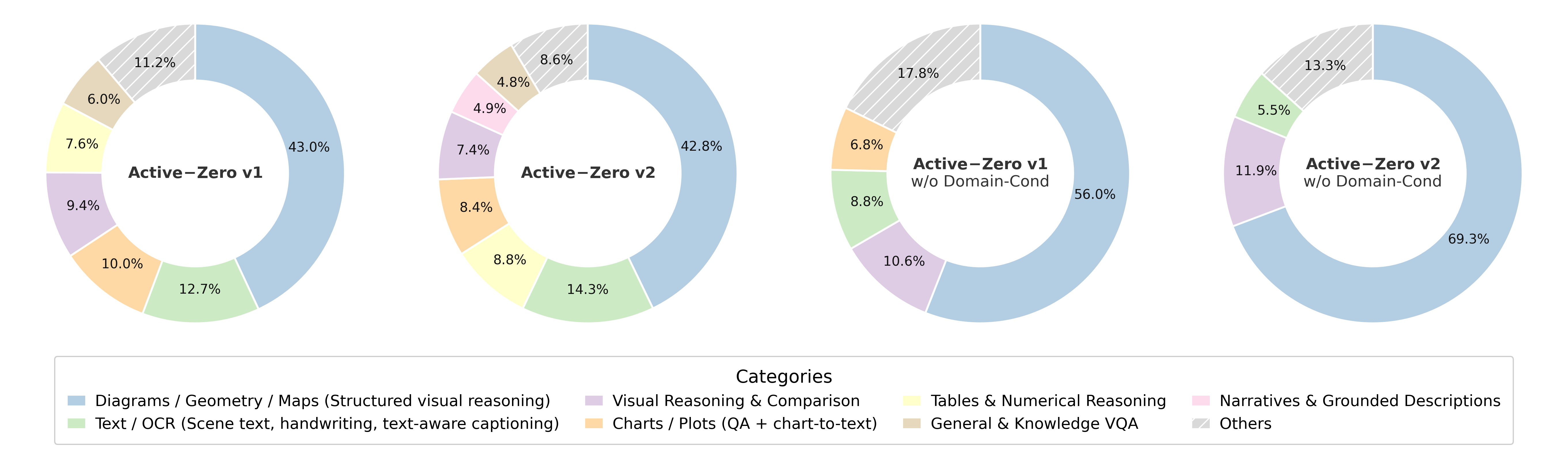}    
	\caption{
    \textbf{Category distribution of Searcher-retrieved images.} Pie charts show the percentage of retrieved images in each task category for Active-Zero v1 and v2, with (left) and without (right) domain conditioning. Categories are identified based on source datasets in $\mathcal{D}_{env}$.
    }
    \label{fig:image_dist}
\end{figure*}

%% file: figs/word_cloud.tex
\begin{figure}[t]
    \centering
    \includegraphics[width=\columnwidth]{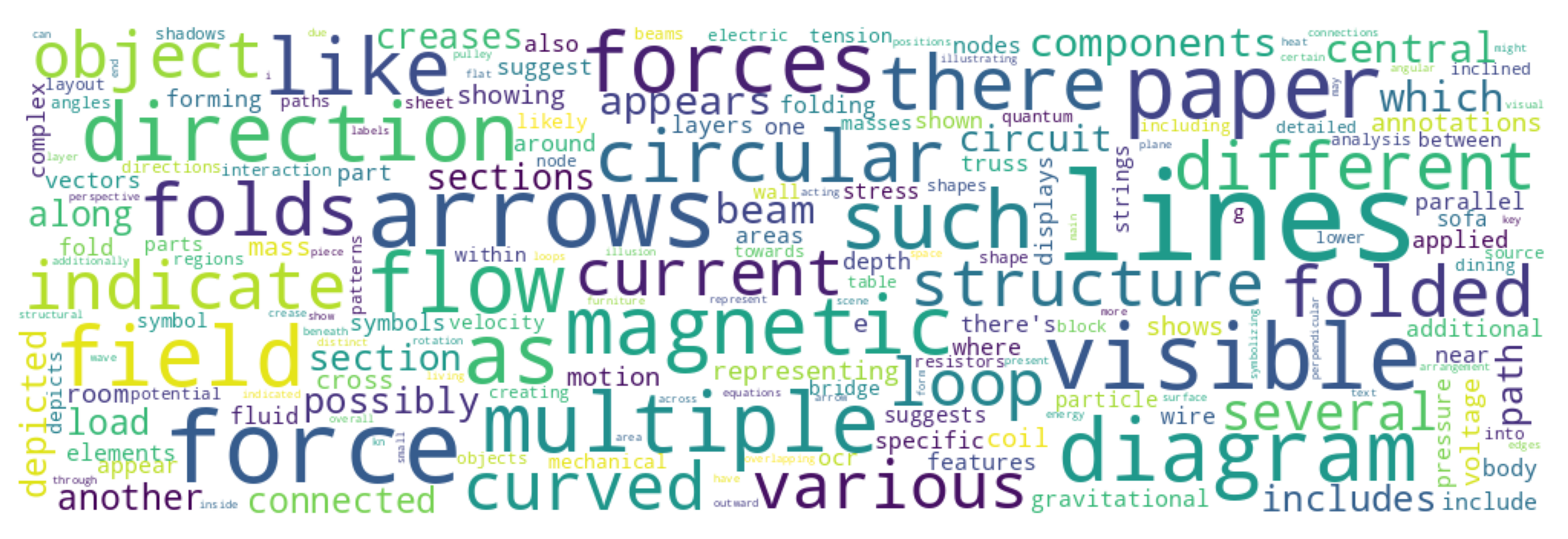}   
	\caption{
    \textbf{Query Distribution Shift Across Searcher Versions.} Word clouds visualizing the query tokens with increased frequency from Searcher v1 to v2. Larger font sizes indicate larger relative frequency gains.
    }
    \label{fig:word_cloud}
\end{figure}

%% file: sec/5_conclusion.tex
\section{Conclusion}

We presented \method{}, a self-play framework that rethinks how vision-language models improve in open-world settings. Rather than passively training on static image collections, \method{} enables models to actively explore visual environments and autonomously construct curricula aligned with their evolving capabilities. By closing the loop between data acquisition, task generation, and reasoning optimization, the proposed Searcher–Questioner–Solver tri-agent system transforms self-play into a dynamic, self-scaffolding learning process. Extensive experiments across 12 benchmarks demonstrate that active environment exploration leads to consistent and complementary gains in both reasoning-intensive tasks and general visual understanding, paving the way for more adaptive VLM reasoning.

%% file: sec/6_appendix.tex
\section{Exeriment Setup}
\label{sec:setup}

\subsection{Evaluation Benchmarks.}
We evaluate models on two groups of benchmarks: (i) reasoning-intensive visual math/logic suites and (ii) general visual understanding + knowledge suites. Below we introduce each benchmark in the order shown in our tables.

\begin{itemize}[leftmargin=0.4cm]
    \item \textbf{MathVista} \cite{lu2023mathvista} is a consolidated benchmark for mathematical reasoning in a visual context. It aggregates problems from many prior multimodal math datasets and also includes newly created subsets, targeting fine-grained perception (reading plots/diagrams/tables) and multi-step reasoning. Evaluation is conducted on testmini split, with 1,000 samples.
    \item \textbf{MathVision (MATH-Vision)} \cite{wang2024measuring} is a curated set of competition-style math problems with visual context, designed to expand subject coverage and difficulty beyond earlier visual-math benchmarks. It spans multiple math disciplines and difficulty levels to stress robust multimodal math reasoning. Evaluation is conducted on the test split, with 3,040 samples.
    \item \textbf{WeMath (WE-MATH)} \cite{qiao2025we} focuses on process-level visual mathematical reasoning. Besides end accuracy, it emphasizes diagnosing how models solve problems by organizing questions along hierarchical knowledge concepts and finer-grained skill layers. Evaluation is conducted on the testmini split, with 1,740 samples.
    \item \textbf{MathVerse} \cite{zhang2024mathverse} is a holistic visual-math benchmark with curated diagram-based problems. A key feature is that each problem can be presented in multiple “versions” with different modality/information availability, enabling controlled analysis of whether models truly use the diagram and how multimodal information contributes to solving. We use the Vision Dominant + Vision Intensive variants filtered from the testmini split, comprising 1,576 samples, which emphasize solving with strong dependence on the image content.  
    \item \textbf{LogicVista} \cite{xiao2024logicvista} targets general logical reasoning in visual contexts (beyond math), with multiple-choice annotated questions spanning several categories of reasoning skills (e.g., inductive/deductive/spatial/numerical/mechanical-style reasoning). Evaluation is conducted on the test split, with 448 samples.
    \item \textbf{DynaMath} \cite{zou2024dynamath} evaluates robustness in multimodal mathematical reasoning under dynamic visual/textual variations. It represents seed questions as programs to automatically generate diverse instantiations, testing stability under perturbations rather than performance on a single static set. Evaluation is conducted on all splits, with 5,010 samples.
    \item \textbf{VisNum (VisNumBench)} \cite{weng2025visnumbench} measures visual number sense: estimating/counting/comparing quantities and other numerical attributes from images, using both synthetic and real-world scenarios to probe how well models ground numeric reasoning in perception. Evaluation is conducted on the test split, with 1,913 samples.
    \item \textbf{RealWorld (RealWorldQA)} \cite{xai_grok15v_2024} is designed to test basic real-world spatial understanding from everyday photos (e.g., relative positions, simple spatial relations) with straightforward, verifiable answers—often easy for humans but challenging for models. Evaluation is conducted on the test split, with 765 samples.
    \item \textbf{MMStar} \cite{chen2024we} is an elite vision-indispensable multimodal benchmark built via careful human selection to reduce spurious shortcuts and ensure visual dependency. It evaluates multiple core capability axes and is commonly used as a general diagnostic of LVLM multimodal competence. Evaluation is conducted on the val split, with 1,500 samples.
    \item \textbf{MMMU} \cite{yue2024mmmu} is a large-scale, multi-discipline benchmark drawn from college-level exams/quizzes/textbooks, requiring subject knowledge and deliberate reasoning over diverse visual formats (e.g., diagrams, tables, charts). It is widely used to assess knowledge-intensive multimodal understanding. Evaluation is conducted on the test set with 895 samples.
    \item \textbf{MMMU\textunderscore{}Pro} \cite{yue2025mmmu} is a more robust variant of MMMU designed to better isolate genuine multimodal understanding (e.g., reducing text-only solvable items and strengthening evaluation settings so models must integrate vision and text). We use the standard (4 options) test split (filtered to single-image questions), comprising 1,591 samples.
    \item \textbf{Hallusion (HallusionBench)} \cite{guan2024hallusionbench} evaluates hallucination and visual illusion failures in vision-language models via expert-written question sets with control-style structure, stressing consistency and grounded image-context reasoning beyond simple object hallucination checks. Evaluation is conducted on the test split, with 951 samples.
\end{itemize}

\subsection{Implementation Details}

In our iterative self-play setup, we conduct three full cycles. Within each iteration, we train the Searcher and Questioner for 10 optimization steps and the Solver for 30 optimization steps. We report results from the second iteration, which yields the best performance. We use a batch size of 512 with a PPO minibatch size of 128. The sampling temperature is 1.0, and the maximum generation length is 2048 tokens. To produce reliable pseudo-labels, we sample $K=9$ trajectories and apply majority voting. To compute uncertainty-based rewards for the Searcher and Questioner, we perform $m=10$ rollouts.

\subsection{Models and baselines}

As the foundational backbone of our framework, we employ the Qwen2.5-VL series \cite{bai2025qwen2}, specifically the 3B-Instruct and 7B-Instruct variants. For baselines, we compare with recent self-play methods developed for VLMs. For VisPlay \cite{he2025visplay}, we reproduce the results closely following the official open-source repository and the training procedures provided by the authors. For Vision-Zero \cite{wang2025vision} and EvoLMM \cite{thawakar2025evolmm}, we directly evaluate the publicly released model weights.

\section{Prompt Templates}
\label{sec:prompt}
\input{prompts/questioner}

\section{Detailed Results}

\input{tables/compl_table1}

\input{tables/compl_table2}

\Cref{tab:compl_table1} and \cref{tab:compl_table2} present a comprehensive comparison of \method{} against strong baselines across both reasoning-intensive and general visual understanding benchmarks, evaluated at two model scales (Qwen2.5-VL-3B-Instruct and Qwen2.5-VL-7B-Instruct) and over multiple self-play iterations.

\textbf{Reasoning-Intensive Benchmarks}
On reasoning-intensive tasks, \method{} consistently improves performance over the base model across all six benchmarks and both model sizes, demonstrating the effectiveness and robustness of the proposed training strategy.
For Qwen2.5-VL-3B-Instruct, the base model achieves an average score of 41.34. After applying \method{}, performance increases substantially, peaking at 45.33 (+3.99) in Iter2. Gains are broad and systematic: MathVista (+2.40), WeMath (+7.70), MathVerse (+4.25), LogicVista (+3.34), and DynaMath (+4.91) all see notable improvements. Iter3 shows a slight regression relative to Iter2, suggesting diminishing returns or mild overfitting with continued iterations at this scale.
For Qwen2.5-VL-7B-Instruct, \method{} yields great and stable improvements. The average score increases from 51.05 (base) to 53.97 (+2.92) at Iter2, with Iter3 remaining competitive at 53.47. The most pronounced gains are observed on LogicVista (43.08 → 48.21, +5.13) and WeMath (64.48 → 69.60, +5.12), highlighting \method{}’s effectiveness on benchmarks that require multi-step logical reasoning and mathematical abstraction. Improvements are also consistent on MathVista, MathVision, and MathVerse, indicating that the method generalizes across diverse reasoning formats rather than targeting a narrow task type.
Overall, these results show that \method{} not only improves average performance but does so uniformly across benchmarks, reinforcing its advantage over static or manually designed curricula.

\textbf{General Visual Understanding and Knowledge-Based Benchmarks}

\Cref{tab:compl_table2} examines whether improvements on reasoning-intensive tasks trade off against general visual understanding or factual knowledge. Overall, \method{} preserves—and often strengthens—performance on these complementary benchmarks.
For Qwen2.5-VL-3B-Instruct, the average score rises from 49.32 to 52.55 (+3.23) at Iter2. The largest gains come from VisNum (+6.80), MMMU (+4.47), and MMMU\textunderscore{}Pro (+4.71), indicating better numerical and multidisciplinary reasoning. While RealWorld drops slightly compared to the base model, improvements on the remaining benchmarks more than offset this decline, yielding a clear net gain.
For Qwen2.5-VL-7B-Instruct, \method{} increases the average from 57.51 to 59.77 (+2.26) at Iter2, with the most pronounced gains on RealWorld (+3.92), VisNum (+3.09), and MMMU (+2.90). Although Iter3 shows mild regressions on MMMU\textunderscore{}Pro and Hallusion relative to Iter2, results remain consistently above the base model, suggesting that additional iterations do not meaningfully erode general capabilities.

\textbf{Iterative Behavior and Scaling Trends}

Across both tables, Iter2 consistently emerges as the strongest iteration, striking a balance between exploration and stability. Iter3 often yields marginal regressions, particularly on smaller models, suggesting that excessive self-play iterations may introduce noise or reduce data diversity. Importantly, the gains from \method{} persist and scale well from 3B to 7B models, demonstrating that the method is not tied to a specific model capacity.

\input{tables/compl_ablation}

%% file: prompts/questioner.tex
\begin{tcolorbox}[
    title=Solver Prompt Template,
    colback=white,
    colframe=gray,
    coltitle=white,
    fonttitle=\bfseries,
    arc=1mm,
    boxrule=0.6mm,
    left=1mm,
    right=1mm,
    top=1mm,
    bottom=1mm,
]

Please reason step by step carefully based on the question: \texttt{\{question\}} and the image.
After completing your reasoning, you MUST output the final, clean, and concise answer strictly inside
\texttt{\textbackslash boxed\{\}}.
The final answer MUST appear inside \texttt{\textbackslash boxed\{\}}, and nowhere else.
If there is no boxed answer, your response is considered incorrect.
\end{tcolorbox}

\begin{tcolorbox}[
    title=Searcher Prompt Template,
    colback=white,
    colframe=gray,
    coltitle=white,
    fonttitle=\bfseries,
    arc=1mm,
    boxrule=0.6mm,
    left=1mm,
    right=1mm,
    top=1mm,
    bottom=1mm,
]
You are an AI Data Engineer. Your task is to generate a unique and concise image retrieval query.\\[2mm]
\textbf{Target Category:} \texttt{\{category\}}\\
\textbf{Focus on:} \texttt{\{query sample\}}\\[1mm]
\textbf{Diversity Requirements:}\\
\textbf{Vary the Subject:} Do not use common examples. Explore different specific objects within the focus area.\\[2mm]
\textbf{Output Format:}\\
\texttt{<type>\{category\}</type>}\\
\texttt{<query>[Concise description emphasizing structure and key markers based on the focus keywords]</query>}\\[2mm]
\textbf{Example (Reference only, DO NOT COPY):}\\
\texttt{<type>\{category\}</type>}\\
\texttt{<query>\{query sample\}</query>}\\
\end{tcolorbox}

\begin{tcolorbox}[
    title=LLM Judge Prompt Template,
    colback=white,
    colframe=gray,
    coltitle=white,
    fonttitle=\bfseries,
    arc=1mm,
    boxrule=0.6mm,
    left=1mm,
    right=1mm,
    top=1mm,
    bottom=1mm,
]
\textbf{System Message:}\\
You are an answer evaluation assistant. Your task is to judge whether two answers are substantially equivalent. When evaluating, you should ignore superficial differences such as format, spaces, punctuation, case, etc., and focus on whether they are consistent in core content, logical meaning and information expression. The judgment criteria should be lenient and inclusive, as long as the expressed meaning is basically the same, it is considered equivalent.\\
\textbf{User Message:}\\
Given the question \texttt{\{question\}}, please judge whether the following two answers express the same meaning. Please only answer "correct" or "incorrect". Correct answer: \texttt{\{ground\_truth\_answer\}}. Answer to be judged: \texttt{\{predicted\_answer\}}. Judgment result (only answer "correct" or "incorrect"). You don't need to reason, just answer "correct" or "incorrect". Don't say anything else. Only answer "correct" or "incorrect" directly without thinking.

\end{tcolorbox}

\begin{tcolorbox}[
    title=Questioner Prompt Template,
    colback=white,
    colframe=gray,
    coltitle=white,
    fonttitle=\bfseries,
    arc=1mm,
    boxrule=0.6mm,
    left=1mm,
    right=1mm,
    top=1mm,
    bottom=1mm,
]
You are an Expert Visual Reasoning Specialist. Your task is to analyze the image, identify core constraints, and generate a high-quality reasoning question that requires multi-step derivation.\\

\textbf{Requirements:}
\vspace{-3mm}
\begin{itemize}
    \item The \textbf{Analysis Phase} must contain:
    \begin{itemize}
        \item \textbf{Visible Elements:} List specific labels, numbers, objects, and spatial markers.
        \item \textbf{Constraints:} Identify explicit or implicit rules (e.g., parallel lines, gravity, total percentages in a chart).
        \item \textbf{Step-by-Step Solution:} Solve your intended question entirely using only the image data to ensure it is mathematically/logically sound.
    \end{itemize}
    \vspace{-2mm}
    \item \textbf{No Simple Description:} Do NOT ask questions that can be answered by simple object identification, color naming, or simple counting.
    \vspace{-2mm}
    \item \textbf{Logical Depth:} The question must require interpreting intent, calculating based on visual cues, spatial transformations, or inferring cause-and-effect.
\end{itemize}

\textbf{Question Type:}\\
1. \texttt{short\_answer}: Requires a short phrase based on logical inference.\\
2. \texttt{numeric\_value}: Requires a calculation or estimation. Provide only the pure number.\\

\textbf{Output Format:}\\
\texttt{<think>}A\texttt{</think>}\\
\texttt{<type>}X\texttt{</type>}\\
\texttt{<question>}Y\texttt{</question>}\\
\texttt{<answer>}Z\texttt{</answer>}\\

\textbf{Strict Rules:}
\vspace{-2mm}
\begin{itemize}
    \item A must be the analysis phase.
    \vspace{-2mm}
    \item X must be one of: \texttt{short\_answer} or \texttt{numeric\_value}.
    \vspace{-2mm}
    \item Z must be the minimal correct value (phrase or number).
    \vspace{-2mm}
    \item Do not add any extra text, labels, or explanations.
\end{itemize}

\vspace{2mm}
\textbf{Examples of Reasoning-based Questions:}\\[1mm]

\texttt{<think>}Visible Elements: A balanced scale shows 3 circles on the left and 2 squares on the right. Constraints: Balanced means equal total weight. Step-by-Step Solution: $3c = 2s \Rightarrow s = 1.5c$, so a square is heavier than a circle.\texttt{</think>}\\
\texttt{<type>}short\_answer\texttt{</type>}\\
\texttt{<question>}Which is heavier: one square or one circle?\texttt{</question>}\\
\texttt{<answer>}square\texttt{</answer>}\\[1mm]

\texttt{<think>}Visible Elements: A triangle with base labeled y, left side labeled 12, right side labeled x, base angles labeled $30^\circ$ (left) and $60^\circ$ (right), and a right-angle marker at the top indicating $90^\circ$. Constraints: A $30^\circ$-$60^\circ$-$90^\circ$ triangle has side ratios $1:\sqrt{3}:2$ (opposite $30^\circ$, $60^\circ$, $90^\circ$ respectively). Step-by-Step Solution: The side 12 is opposite $60^\circ$, so $12=\sqrt{3}k \Rightarrow k=\frac{12}{\sqrt{3}}=4\sqrt{3}$. The hypotenuse is $y=2k=8\sqrt{3}$.\texttt{</think>}\\
\texttt{<type>}numeric\_value\texttt{</type>}\\
\texttt{<question>}Find y.\texttt{</question>}\\
\texttt{<answer>}$8\sqrt{3}$\texttt{</answer>}\\

\vspace{2mm}
\vspace{2mm}
\end{tcolorbox}

%% file: tables/compl_table1.tex
\begin{table*}[t!]
\centering
\caption{Comparison of \method{} against state-of-the-art baselines on reasoning-intensive benchmarks.}
\label{tab:compl_table1}
\small
\resizebox{\textwidth}{!}{
\begin{tabular}{p{3cm}p{1.6cm}p{1.6cm}p{1.6cm}p{1.6cm}p{1.6cm}p{1.6cm}p{1.6cm}p{1.6cm}}
\toprule[1.5pt]
\textbf{Method} & \textbf{MathVista} & \textbf{MathVision} & \textbf{WeMath} & \textbf{MathVerse} & \textbf{LogicVista} & \textbf{DynaMath} & \textbf{Average}\\
\midrule
\midrule
\multicolumn{9}{@{}l}{\textit{Qwen2.5-VL-3B-Instruct}} \\
\cdashline{1-8}[2pt/2pt]
\rule{0pt}{10pt}%
 Base Model               & 57.80 & 22.27 & 53.51 & 30.84 & 37.28 & 46.33 & 41.34 \\
\cdashline{1-8}[2pt/2pt]
\rule{0pt}{10pt}%
\method{} (Iter1)               & 59.20 & 23.06 & 60.06 & 33.88 & 39.96 & 50.30 & 44.41 \\
 \method{} (Iter2)              & 60.20 & 23.62 & 61.21 & 35.09 & 40.62 & 51.24 & 45.33\\
 \method{} (Iter3)              & 59.00 & 22.20 & 61.09 & 33.88 & 40.40 & 51.00 & 44.60\\
\midrule
\multicolumn{9}{@{}l}{\textit{Qwen2.5-VL-7B-Instruct}} \\
\cdashline{1-8}[2pt/2pt]
\rule{0pt}{10pt}%
 Base Model               & 69.40 & 25.95 & 64.48 & 45.11 & 43.08 & 58.26 & 51.05 \\
\cdashline{1-8}[2pt/2pt]
\rule{0pt}{10pt}%
 \method{} (Iter1)              & 71.60 & 26.97 & 68.16 & 46.76 & 46.21 & 59.54 & 53.21 \\
 \method{} (Iter2)              & 72.60 & 26.18 & 69.60 & 48.16 & 48.21 & 59.04 & 53.97 \\
 \method{} (Iter3)              & 72.30 & 26.61 & 69.94 & 45.88 & 47.54 & 58.52 & 53.47 \\
\bottomrule[1.5pt]
\end{tabular}}
\end{table*}

%% file: tables/compl_table2.tex
\begin{table*}[t!]
\centering
\caption{Comparison of \method{} against state-of-the-art baselines on general visual understanding and knowledge-based benchmarks.}
\label{tab:compl_table2}
\small
\resizebox{\textwidth}{!}{
\begin{tabular}{p{3cm}p{1.6cm}p{1.6cm}p{1.6cm}p{1.6cm}p{1.6cm}p{1.6cm}p{1.6cm}p{1.6cm}}
\toprule[1.5pt]
\textbf{Method} & \textbf{VisNum} & \textbf{RealWorld} & \textbf{MMStar} & \textbf{MMMU} & \textbf{MMMU\textunderscore{}Pro} & \textbf{Hallusion} & \textbf{Averge}\\
\midrule
\midrule
\multicolumn{9}{@{}l}{\textit{Qwen2.5-VL-3B-Instruct}} \\
\cdashline{1-8}[2pt/2pt]
\rule{0pt}{10pt}%
 Base Model                        & 34.71 & \textbf{61.70} & 52.80 & 44.69 & 41.99 & 60.04 & 49.32 \\
\cdashline{1-8}[2pt/2pt]
\rule{0pt}{10pt}%
 \method{} (Iter1)                & 40.93 & 61.44 & 53.20 & 47.93 & 46.64 & 64.25 & 52.40  \\
 \method{} (Iter2)               & 41.51 & 60.00 & 53.80 & 49.16 & 46.70 & 64.14 & 52.55 \\
 \method{} (Iter3)               &  41.51 & 60.26 & 52.80 & 48.27 & 45.69 & 63.72 & 52.04 \\
\midrule
\multicolumn{9}{@{}l}{\textit{Qwen2.5-VL-7B-Instruct}} \\
\cdashline{1-8}[2pt/2pt]
\rule{0pt}{10pt}%
 Base Model               & 44.85 & 63.66 & 61.53 & 53.52 & 52.61 & 68.87 & 57.51 \\
\cdashline{1-8}[2pt/2pt]
\rule{0pt}{10pt}%
 \method{} (Iter1)                & 45.53 & 66.01 & 62.40 & 54.30 & 53.30 & 69.61 & 58.53  \\
 \method{} (Iter2)               &  47.94 & 67.58 & 64.00 & 56.42 & 54.62 & 68.03 & 59.77 \\
 \method{} (Iter3)               &  48.35 & 67.45 & 64.33 & 54.75 & 51.60 & 69.51 & 59.33 \\
\bottomrule[1.5pt]
\end{tabular}}
\end{table*}

%% file: tables/compl_ablation.tex
\begin{table*}[t!]
\centering
\caption{Ablation Study of \method{} on reasoning-intensive benchmarks.}
\label{tab:compl_abla1}
\small
\resizebox{\textwidth}{!}{
\begin{tabular}{p{3cm}p{1.6cm}p{1.6cm}p{1.6cm}p{1.6cm}p{1.6cm}p{1.6cm}p{1.6cm}p{1.6cm}}
\toprule[1.5pt]
\textbf{Method} & \textbf{MathVista} & \textbf{MathVision} & \textbf{WeMath} & \textbf{MathVerse} & \textbf{LogicVista} & \textbf{DynaMath} & \textbf{Average}\\
\midrule
\midrule
\multicolumn{9}{@{}l}{\textit{Qwen2.5-VL-3B-Instruct}} \\
\cdashline{1-8}[2pt/2pt]
\rule{0pt}{10pt}%
 Base Model               & 57.80 & 22.27 & 53.51 & 30.84 & 37.28 & 46.33 & 41.34 \\
\cdashline{1-8}[2pt/2pt]
\rule{0pt}{10pt}%
\method{}                    & 60.20 & 23.62 & 61.21 & 35.09 & 40.62 & 51.24 & 45.33 \\
Base Chanllenger             & 59.30 & 21.88 & 61.32 & 34.39 & 40.40 & 51.18 & 44.75 \\
Random Sampling              & 58.60 & 23.16  & 56.72 & 34.96 & 39.06 & 50.80 & 43.88 \\
w/o Domain-Condition         & 59.20  & 22.43  & 62.41  & 35.15 & 39.73 & 50.68 & 44.93 \\
w/o Visual Penalty           & 59.40 & 22.20 & 62.24  & 33.88 & 40.18& 50.72 & 44.89\\
\midrule
\end{tabular}}
\end{table*}

\begin{table*}[t!]
\centering
\caption{Ablation Study of \method{} on general visual understanding and knowledge-based benchmarks.}
\label{tab:compl_abla2}
\small
\resizebox{\textwidth}{!}{
\begin{tabular}{p{3cm}p{1.6cm}p{1.6cm}p{1.6cm}p{1.6cm}p{1.6cm}p{1.6cm}p{1.6cm}p{1.6cm}}
\toprule[1.5pt]
\textbf{Method} & \textbf{VisNum} & \textbf{RealWorld} & \textbf{MMStar} & \textbf{MMMU} & \textbf{MMMU\textunderscore{}Pro} & \textbf{Hallusion} & \textbf{Averge}\\
\midrule
\midrule
\multicolumn{9}{@{}l}{\textit{Qwen2.5-VL-3B-Instruct}} \\
\cdashline{1-8}[2pt/2pt]
\rule{0pt}{10pt}%
 Base Model                    & 34.71 & 61.70 & 52.80 & 44.69 & 41.99 & 60.04 & 49.32 \\
\cdashline{1-8}[2pt/2pt]
\rule{0pt}{10pt}%
\method{}                      & 41.51 & 60.00 & 53.80 & 49.16 & 46.70 & 64.14 & 52.55  \\
Base Challenger                & 40.51 & 57.91 & 52.53 & 47.37 & 44.75 & 63.09 & 51.03  \\
Random Sampling                & 37.69 & 63.53 & 53.27 & 47.82 & 44.06 & 64.77 & 51.86  \\
w/o Domain-Condition           & 40.04 & 57.39 & 52.20 & 47.71 & 45.13 & 61.93 & 50.73  \\
w/o Visual Penalty             & 41.82 & 59.61 & 54.13 & 47.15 & 44.88 & 62.15 & 51.62  \\
 \bottomrule
\end{tabular}}
\end{table*}